\documentclass{fairmeta}

\usepackage{amsmath,amssymb,amsfonts,amsthm}
\usepackage{bbm}
\usepackage{tabularx}
\usepackage{array}
\usepackage{enumitem}
\usepackage{pifont}
\usepackage{tikz}
\usepackage{silence}
\usepackage{booktabs}
\usepackage{multirow}
\usepackage{xspace}
\usepackage{longtable}

\usetikzlibrary{arrows.meta,positioning}

\setlist[itemize]{topsep=3pt, itemsep=2pt, parsep=0pt, partopsep=0pt}
\setlist[enumerate]{topsep=3pt, itemsep=2pt, parsep=0pt, partopsep=0pt}

\newcommand{\benchmark}{\textsc{TxnQA}\xspace}

\title{\benchmark: Benchmarking Analytical Reasoning and Evidence Grounding over Transaction Event Sequences}

\author[1]{Rabimba Karanjai}
\author[1]{Yiming Bian}
\author[1]{Yun-Shiuan Chuang}
\author[1,2,*]{Jenna L. Ballard}
\author[1]{Zijie Pan}
\author[1]{Shivani Shekhar}
\author[1]{Yixi Li}
\author[1]{Yirou Ge}
\author[1]{Avinash Thangali}
\author[1]{Richard Williamson}
\author[1]{Linsey Pang}
\author[1]{Prakhar Mehrotra}

\affiliation[1]{PayPal AI}
\affiliation[2]{University of Pennsylvania}
\contribution[*]{Work done during internship at PayPal.}
\correspondence{rkaranjai@paypal.com}
\date{October 2026}

\abstract{Analyzing merchant transaction histories requires dynamic temporal filtering, entity resolution, and aggregation over irregular event streams---operations poorly captured by static tabular or time-series forecasting benchmarks. We introduce \benchmark, an analytical QA benchmark comprising 27 task families (81 templates) alongside a 5-scenario fraud-risk reasoning suite that evaluates both verdicts and supporting evidence over synthetic, privacy-mitigated transaction data. Ground-truth answers are computed programmatically, and model performance is decomposed into format-following rate (FFR), conditional content accuracy (CA), and end-to-end correctness (E2E). Evaluating prompted proprietary baselines and six LoRA-adapted open-weight models shows that adaptation reliably enforces output schemas but leaves arithmetic bottlenecks: Gemma-4-31B-IT and Qwen3-8B attain E2E scores of 0.867 and 0.816, whereas evaluated MoE checkpoints reach only 0.632--0.713. Furthermore, on the 500-instance fraud suite, verdict accuracy frequently exceeds $80\%$, yet joint accuracy (requiring exact evidence attribution) collapses to at most ${\sim}23\%$. These findings identify schema compliance, semantic calculation, and evidence attribution as three distinct bottlenecks in transaction-sequence QA.}

\begin{document}

\maketitle

\section{Introduction}
\label{sec:intro}

Natural language interfaces for transaction histories allow analysts to query financial activity directly, determining how many unique buyers made purchases in a given window, identifying peak transaction periods, measuring periods of merchant inactivity, or checking whether median spend shifted following a specific date. While such queries are straightforward to articulate, answering them requires orchestrating multiple operations over an irregular stream of events. A large language model (LLM) must filter records by time, resolve counterparties, execute aggregations or comparisons, and format the final answer to meet downstream operational requirements. In high-stakes applications such as fraud detection, the LLM must also move beyond a binary classification to isolate the specific transactions and attributes that support its conclusion.

A transaction history differs fundamentally from standard tabular data or regularly sampled time series. It pairs static merchant metadata with an irregular sequence of timestamped events featuring mixed numerical and categorical attributes. Furthermore, the required computation depends entirely on the query: one question might demand an extremum search, another an entity-linkage check, and a third a multi-step aggregation over sliding windows. Rather than a static forecasting problem, transaction question answering (QA) is an analytical task conditioned entirely on the natural language query. While event-sequence modeling has recognized the distinct nature of irregular event data~\citep{li2025survey,sun2026review}, language-driven analytical interfaces over these streams remain underexplored.

Prior work has addressed adjacent problems, including QA over financial reports and tables~\citep{chen2021finqa,zhu2021tatqa,chen2022convfinqa,zhao2022multihiertt}, time-series QA~\citep{kong2025timemqa,jing2026tsaqa,xie2024chatts,han2026tsskill,chen2026irtstoolbench,nie2026clirbench,xie2026arfbench}, representation learning for tabular sequences~\citep{padhi2021tabformer,babaev2022coles,moskvoretskii2024mlem,mollaev2025multimodal}, and safety or fraud evaluation for language models~\citep{yang2025fraudr1,chakraborty2024detoxbench}. However, current benchmarks rarely evaluate whether models can simultaneously reason across irregular event histories, comply with typed answer schemas, and provide grounded, feature-level evidence.

We make four contributions:
\begin{enumerate}[leftmargin=*]
  \item A benchmark containing 27 analytical task families spanning 43 answer-format variants and 5 representative fraud scenarios;
  \item Programmatically generated analytical gold answers and a scheme that scores format compliance (FFR) separately from content correctness (CA) and jointly (E2E);
  \item Prompted-only evaluation across proprietary and open-weight systems and a base-versus-LoRA study across six open-weight checkpoints, compared against a time-series-specialized model;
  \item Evidence that nominal parameter scale is not monotonic with adapted performance, and that fraud-verdict quality can substantially exceed supporting-evidence quality.
\end{enumerate}

\section{Related Work}
\label{sec:related-work}

\paragraph{Financial Question Answering.}
Prior financial QA benchmarks primarily target semi-structured tabular records and narrative text in financial reports. Notably, FinQA evaluated numerical reasoning via executable programs~\citep{chen2021finqa}, TAT-QA combined hybrid tabular and textual contexts~\citep{zhu2021tatqa}, ConvFinQA introduced multi-turn conversational financial reasoning~\citep{chen2022convfinqa}, and MultiHiertt addressed hierarchical, multi-page document structures~\citep{zhao2022multihiertt}. While these works underscore the necessity of exact numerical evaluation and structured derivation traces, their underlying contexts are static documents and tables rather than dynamic event streams. Consequently, they do not evaluate temporal dynamics unique to transaction histories, such as sliding-window aggregations, chronological boundary filtering, or cross-transaction entity resolution.

\paragraph{Time-Series and Event-Sequence QA.}
A growing body of work examines language-model interfaces for temporal data. Time-MQA established a broad multi-task benchmark for continual pre-training on regular time series~\citep{kong2025timemqa}, while ChatTS unified multivariate time series as a native input modality~\citep{xie2024chatts}. Parallel efforts have aligned time-series embeddings with language representations~\citep{sun2024test,langer2025opentslm} and evaluated specialized skills such as tool execution, long-context retrieval, and clinical grounding~\citep{han2026tsskill,chen2026irtstoolbench,nie2026clirbench,xie2026arfbench}. Most relevant to our data format is ESQA~\citep{abdullaeva2026esqa}, which categorizes event-sequence questions into predictive tasks (forecasting future attributes such as loan default) and extractive tasks (computing sequence-level descriptive statistics). Similarly, recent efforts have explored event-centric multi-hop reasoning~\citep{li2024meqa} and financial event foundation models~\citep{rusakov2026financialfm}. \benchmark aligns with the extractive and analytical paradigm of these benchmarks, but differentiates itself by introducing: (i) strictly typed operational schemas that decouple formatting compliance from numerical execution, and (ii) a controlled, feature-level evidence-selection task for risk assessment.

\paragraph{Transaction and Event-Sequence Modeling.}
Dedicated architectures for tabular sequences such as TabFormer~\citep{padhi2021tabformer}, CoLES~\citep{babaev2022coles}, and MLEM~\citep{moskvoretskii2024mlem} use field-aware attention and contrastive objectives to model multi-attribute event histories. While these approaches produce powerful sequence representations, they are predominantly evaluated on fixed downstream objectives such as classification, representation retrieval, or next-token forecasting. \benchmark instead benchmarks language models on query-conditioned analytical reasoning, where the required operation must be interpreted dynamically from an analyst's query.

\paragraph{Fraud QA Benchmarking.}
Prior LLM fraud benchmarks mainly focus on textual content. Fraud-R1~\citep{yang2025fraudr1} evaluates resistance to multi-turn fraud inducements, while DetoxBench~\citep{chakraborty2024detoxbench} benchmarks classification across phishing, fraudulent and spam emails, fake jobs, fake news, and abusive language. These settings assess signals expressed directly in text rather than evidence distributed across structured transaction records. Closer to our setting, \citet{tan2026structuredfinancialdata} apply LLMs to tabular transactions and generate rationales alongside fraud scores, but do not quantitatively evaluate whether those rationales identify the correct evidence. \benchmark instead evaluates both verdict accuracy and feature-level evidence selection using closely matched profiles with different outcomes.

\section{Benchmark Construction and Validation}
\label{sec:construction}

\begin{figure*}[t]
\centering
\includegraphics[width=0.72\textwidth]{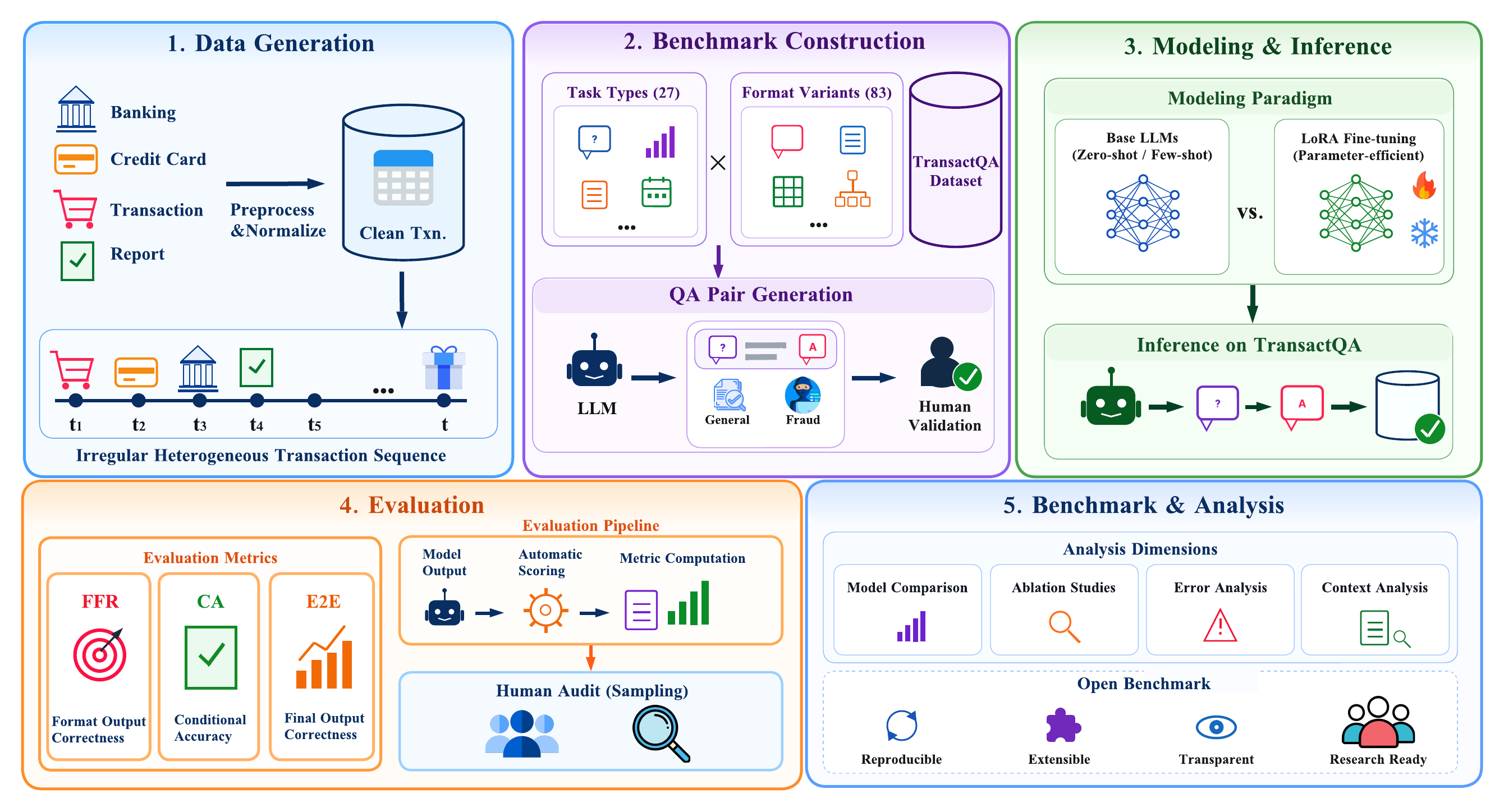}
\caption{Overview of the TxnQA benchmark framework. Starting from real merchant transaction data, we generate privacy-mitigated synthetic merchant profiles and irregular transaction sequences using REaLTabFormer. Each transaction sequence is converted into natural-language context and paired with automatically generated question-answer pairs. The resulting benchmark is used to evaluate multiple open-weight large language models under both base (zero-shot) and LoRA supervised fine-tuning settings. Model performance is assessed using three complementary metrics together with fine-grained analyses across question categories, answer formats, and model architectures.}
\label{fig:overview}
\end{figure*}

\subsection{Transaction Sequence Generation}

Publicly accessible merchant transaction records are scarce due to commercial confidentiality and data privacy constraints. To address this, we train REaLTabFormer~\citep{solatorio2023realtabformer} on approved internal transaction records from a large payment processing platform; because the raw source data cannot be shared, we release only the synthetic derivative. REaLTabFormer structures merchant profiles as a primary (`parent') table and each merchant's transaction sequence as an associated (`child') table. The model autoregressively samples the parent metadata before using a sequence-to-sequence model to generate the corresponding transaction history, dynamically determining sequence lengths.

We generate separate user sets for training, validation, and testing (30{,}000 / 6{,}500 / 180 merchants). The test set contains 162 General QA items and 500 Fraud QA items (Table~\ref{tab:splits}), chosen to cover all 27 analytical task types (6 instances each) and the 5 fraud scenarios. The General QA test set draws its 162 items from 160 distinct merchants; a small number of merchants contribute more than one evaluation item so that every task type reaches its target count.

\begin{table}[t]
\centering
\small
\setlength{\tabcolsep}{6.0pt}
\begin{tabular}{lrrl}
\toprule
Split & Users & Items & Use \\
\midrule
Train & 30,000 & -- & LoRA SFT \\
Validation & 6,500 & -- & Model selection \\
Test -- General QA & 160 & 162 & Main analytical results \\
Test -- Fraud QA & 500 & 500 & Fraud-evidence results \\
\bottomrule
\end{tabular}
\caption{Dataset populations used in the evaluation.}
\label{tab:splits}
\end{table}

\paragraph{Post-processing and quality assurance.}
To preserve logical consistency, we re-compute all deterministic fields directly from their foundational attributes and re-index timestamps to guarantee strict chronological ordering. We design empirical diagnostics along three dimensions: statistical realism (Kolmogorov--Smirnov tests for continuous features, total variation distance for discrete distributions), relational fidelity (conservation of Spearman rank correlations across parent and child tables), and sample novelty (exact-match detection and distance-to-closest-record against held-out private records). Because these metrics represent empirical diagnostics rather than formal differential privacy guarantees, we classify the released benchmark as \textit{privacy-mitigated}.

\paragraph{Transaction text generation.}
Each structured sequence is serialized into natural language by combining a summary sentence of the merchant profile with chronological transaction descriptions (e.g., \textit{``On Monday, March 3, 2025 at 12:15pm, a transaction of \$24.50 was processed.''} ). This textual narrative forms the sole context provided to the language model for all queries concerning that merchant.

\paragraph{QA generation.}
We define 27 analytical question types, each with 3 wording variants (81 templates total) and 43 answer-format variants spanning precise and categorical regimes (full enumeration in Appendix~\ref{app:questions}); gold answers are computed directly from the structured records using the formula appropriate to each question type, not generated or paraphrased by a model. Three question types are not answerable from the rendered text alone and have gold answer \textsc{unknown}, scored via a decline-accuracy metric. The FraudQA set instantiates five fraud scenarios---merchant batch signup under a gibberish email domain, unusual-amount transfer between the same entity, impossible-travel IP change, device-spoof compromise, and card testing (full descriptions in Appendix~\ref{app:questions})---each programmatically expanded into $100$ cases balanced across confirmed-fraud, hard-negative (benign look-alike), and clear-negative instances, with reference verdicts and gold evidence features specified by the scenario design.

\section{Models and Training}
\label{sec:models}

To provide a comprehensive evaluation across model scales, architectures, and reasoning paradigms, we benchmark 13 open-weight checkpoints alongside three leading proprietary systems.

Our open-weight selection spans dense and mixture-of-experts (MoE) architectures across scales from 7B to 235B parameters: DeepSeek-R1-Distill-Qwen-7B, Qwen3-8B, Qwen3-32B, Qwen3-30B-A3B-Instruct-2507, Qwen3-30B-A3B-Thinking-2507, Qwen3-235B, Gemma-4-31B-IT, GLM-5.2, GPT-OSS-20B, GPT-OSS-120B, DeepSeek-V4-Flash, DeepSeek-V4-Pro, and Kimi-K2.6~\citep{guo2025deepseek,yang2025qwen3,team2503gemma,zeng2026glm,agarwal2025gpt,xu2026deepseek,team2026kimi}. For models supporting configurable test-time compute, we evaluate both reasoning-enabled (``thinking'') and direct generation modes. For proprietary baselines, we evaluate current frontier models---Claude Sonnet, GPT-5.2, and Gemini 2.5 Pro---under zero-shot prompting via standard APIs. Finally, we evaluate ChatTS~\citep{xie2024chatts} as an architectural contrast that encodes transaction histories directly as numerical time series rather than serialized text.

We adapt six representative open-weight checkpoints using Low-Rank Adaptation (LoRA; \citealp{hu2021lora}): DeepSeek-R1-Distill-7B, Qwen3-8B, Qwen3-32B, Qwen3-30B-A3B-Instruct, Qwen3-30B-A3B-Thinking, and Gemma-4-31B-IT. We apply adapters to all attention and feed-forward projections with rank $r=32$, scaling factor $\alpha=64$, and dropout rates of $0.00$ or $0.05$. Training runs for 2--3 epochs with learning rates tuned between $5\times10^{-5}$ and $1\times10^{-4}$ using cosine decay (detailed hyperparameters are listed in Appendix~\ref{app:hyperparams}).

We note an important experimental distinction regarding the two Qwen3-30B-A3B variants: although they share identical parameter counts (30B total, 3B active), they represent distinct post-training recipes rather than an isolated toggle on reasoning compute. Furthermore, our fine-tuning recipe adapts the Thinking variant using a direct-answer instruction template rather than eliciting intermediate chain-of-thought traces, an operational choice whose implications we analyze in Section~\ref{sec:discussion}.

\section{Evaluation Protocol}
\label{sec:protocol}

We evaluate all systems across the 162 General QA and 500 Fraud QA test instances. To account for stochasticity during generation, each model produces three independent completions per question (random seeds 0, 1, and 2) using temperature-based sampling ($T=0.7$). We report the empirical mean and standard deviation across these three decoding runs.

For statistical comparisons across checkpoints, significance markers denote Welch's one-tailed two-sample $t$-tests with Bonferroni correction ($\alpha = 0.05$) applied across seed aggregates. Open-weight models were evaluated on NVIDIA B200 GPUs (Ubuntu 24.04, CUDA 12.9, PyTorch 2.10, vLLM 0.19.1), while proprietary models were queried via standardized API endpoints.

\paragraph{Metrics.}
To isolate whether errors stem from syntax non-compliance or flawed arithmetic, let $F_j \in \{0, 1\}$ indicate format validity and $C_j \in \{0, 1\}$ indicate normalized correctness for example $j \in \{1, \dots, N\}$. We define:
\begin{equation}
\label{eq:e2e}
\mathrm{FFR} = \frac{1}{N}\sum_j F_j, \quad
\mathrm{CA} = \frac{\sum_j F_j C_j}{\sum_j F_j}, \quad
\mathrm{E2E} = \frac{1}{N}\sum_j F_j C_j.
\end{equation}
Within any run, $\mathrm{E2E} = \mathrm{FFR} \times \mathrm{CA}$. In multi-seed tables, we report $\mu_{\mathrm{FFR}} \times \mu_{\mathrm{CA}}$, which closely approximates mean per-run E2E.

For Fraud QA, an answer contains a verdict and cited feature set $P$; let $G$ denote gold evidence and $I = |P \cap G|$. We report: (1)~\textbf{FFR}: valid (verdict, evidence) parses; (2)~\textbf{Verdict Accuracy}: accuracy among valid outputs; and (3)~\textbf{Joint Accuracy}: strict exact-match ($P = G$ and correct verdict). To characterize failure modes without diluting signal, evidence quality is macro-averaged exclusively over rows missing exact evidence ($P \neq G$):
\begin{itemize}[leftmargin=*]
  \item \textbf{Recall} $= I / |G|$: share of required gold features recovered;
  \item \textbf{Precision} $= I / |P|$: share of cited features that are correct;
  \item \textbf{Over-citation} $= |P \setminus G| / |G|$: number of extraneous cited features as a multiple of gold-set size (lower is better).
\end{itemize}

\section{Results and Diagnostic Analysis}
\label{sec:results}

\subsection{Zero-Shot Baselines}

Table~\ref{tab:proprietary-zeroshot} summarizes zero-shot performance across proprietary systems, while Appendix Table~\ref{tab:opensource-zeroshot} provides complete results across open-weight checkpoints.

Among proprietary systems, GPT-5.2 (high reasoning) attains the highest strict content accuracy on precise questions (88.7\% CA), whereas Claude Sonnet (thinking enabled) achieves the strongest output schema adherence (97.5\% precise FFR, 99.1\% categorical FFR) and the highest overall end-to-end score (82.9\% E2E). Across providers, enabling test-time reasoning yields mixed effects rather than uniform gains: while it consistently improves strict numerical accuracy for GPT-5.2 (+12.4\% CA) and Claude Sonnet (+0.0\% CA, but +12.4\% FFR), its impact on Gemini 2.5 Pro is marginal across both precise and categorical tasks. These variations indicate that extended reasoning traces do not inherently guarantee better schema compliance without explicit output constraints.

\begin{table*}[t]
\centering
\small
\setlength{\tabcolsep}{6.0pt}
\begin{tabular}{llccccc}
\toprule
Proprietary model & Reasoning & \multicolumn{3}{c}{Precise QA} & \multicolumn{2}{c}{Categorical QA} \\
\cmidrule(lr){3-5} \cmidrule(lr){6-7}
 & & FFR & CA (strict) & E2E (approx.) & FFR & CA (match) \\
\midrule
Claude Sonnet & Thinking-on  & \textbf{97.5 $\pm$ 1.1} & 85.0 $\pm$ 0.5 & \textbf{82.9 $\pm$ 1.1} & \textbf{99.1 $\pm$ 1.6} & \textbf{74.8 $\pm$ 0.4} \\
Claude Sonnet & Thinking-off & 85.1 $\pm$ 1.5 & 85.1 $\pm$ 2.3 & 72.4 $\pm$ 2.3 & 97.2 $\pm$ 0.0 & 70.5 $\pm$ 1.6 \\
GPT-5.2 & High  & 87.3 $\pm$ 0.5 & \textbf{88.7 $\pm$ 0.6} & 77.4 $\pm$ 0.7 & 95.4 $\pm$ 1.6 & 73.8 $\pm$ 3.0 \\
GPT-5.2 & None & 89.8 $\pm$ 0.5 & 76.3 $\pm$ 2.5 & 68.5 $\pm$ 2.3 & 88.9 $\pm$ 2.8 & 70.8 $\pm$ 2.0 \\
Gemini 2.5 Pro & High  & 91.4 $\pm$ 1.6 & 81.2 $\pm$ 0.3 & 74.2 $\pm$ 1.3 & 91.7 $\pm$ 0.0 & 63.6 $\pm$ 3.0 \\
Gemini 2.5 Pro & Low & 89.8 $\pm$ 2.2 & 82.7 $\pm$ 1.9 & 74.3 $\pm$ 2.5 & 93.5 $\pm$ 4.2 & 67.4 $\pm$ 1.5 \\
\bottomrule
\end{tabular}
\caption{Zero-shot performance of proprietary models, percentages with SD across 3 decoding seeds. E2E is approximated as $\mu_{FFR}\times\mu_{CA}/100$; SD via delta-method approximation. Bold indicates top value per column.}
\label{tab:proprietary-zeroshot}
\end{table*}

\subsection{Fine-Tuned Performance and Base-to-Tuned Gains}

Table~\ref{tab:mainresults} presents performance following LoRA adaptation on open-weight models, alongside the specialized time-series baseline ChatTS. Table~\ref{tab:gains} isolates the specific gains from base checkpoints to adapted variants under matched protocols.

\begin{table*}[t]
\centering
\small
\setlength{\tabcolsep}{3.5pt}
\begin{tabular}{lcccccc}
\toprule
Fine-tuned checkpoint & \multicolumn{3}{c}{Precise QA} & \multicolumn{2}{c}{Categorical QA} & Decline Acc. \\
\cmidrule(lr){2-4} \cmidrule(lr){5-6}
 & FFR & CA (strict) & E2E (approx.) & FFR & CA (match) \\
\midrule
DeepSeek-R1-Distill-7B      & \textbf{100.0 $\pm$ 0.0} & 79.0 $\pm$ 2.5 & 79.0 $\pm$ 2.5 & \textbf{100.0 $\pm$ 0.0} & \textbf{91.7 $\pm$ 2.8} & \textbf{100.0} \\
Qwen3-8B                    & \textbf{100.0 $\pm$ 0.0} & 81.6 $\pm$ 2.4 & 81.6 $\pm$ 2.4 & \textbf{100.0 $\pm$ 0.0} & \textbf{91.7 $\pm$ 2.8} & \textbf{100.0} \\
Qwen3-30B-A3B-Instruct-2507 & 99.0 $\pm$ 0.0           & 72.1 $\pm$ 1.0 & 71.3 $\pm$ 1.0 & \textbf{100.0 $\pm$ 0.0} & 88.0 $\pm$ 1.6 & \textbf{100.0} \\
Qwen3-30B-A3B-Thinking-2507 & \textbf{100.0 $\pm$ 0.0} & 63.2 $\pm$ 1.1 & 63.2 $\pm$ 1.1 & \textbf{100.0 $\pm$ 0.0} & 76.9 $\pm$ 3.2 & \textbf{100.0} \\
Qwen3-32B                   & \textbf{100.0 $\pm$ 0.0} & 82.5 $\pm$ 3.1 & 82.5 $\pm$ 3.1 & \textbf{100.0 $\pm$ 0.0} & 89.8 $\pm$ 1.6 & \textbf{100.0} \\
Gemma-4-31B-IT              & \textbf{100.0 $\pm$ 0.0} & 86.7 $\pm$ 2.5 & 86.7 $\pm$ 2.5 & \textbf{100.0 $\pm$ 0.0} & 88.0 $\pm$ 1.6 & \textbf{100.0} \\
ChatTS$^\dagger$            & \textbf{100.0} & \textbf{87.6} & \textbf{87.6} & \textbf{100.0} & 88.9 & \textbf{100.0} \\
\bottomrule
\end{tabular}
\caption{Fine-tuned performance on the General QA precise- and categorical-answer splits. E2E approximated as in Table~\ref{tab:proprietary-zeroshot}. $^\dagger$ChatTS evaluated with a single deterministic run.}
\label{tab:mainresults}
\end{table*}

\begin{table*}[t]
\centering
\small
\setlength{\tabcolsep}{2.2pt}
\footnotesize\begin{tabular}{llrrrrrr}
\toprule
Protocol & Checkpoint & Base FFR & Tuned FFR & $\Delta$FFR & Base E2E & Tuned E2E & $\Delta$E2E \\
\midrule
\multirow{4}{*}{Matched non-thinking}
& Qwen3-8B & 0.981 $\pm$ 0.00 & 1.000 $\pm$ 0.00 & +0.019 & 0.505 $\pm$ 0.02 & 0.816 $\pm$ 0.02 & +0.311 \\
& Gemma-4-31B-IT & 0.956 $\pm$ 0.01 & 1.000 $\pm$ 0.00 & +0.044 & 0.711 $\pm$ 0.01 & 0.867 $\pm$ 0.03 & +0.156 \\
& Qwen3-32B & 0.978 $\pm$ 0.01 & 1.000 $\pm$ 0.00 & +0.022 & 0.555 $\pm$ 0.03 & 0.825 $\pm$ 0.03 & +0.270 \\
& Qwen3-30B-A3B-Instruct & 0.959 $\pm$ 0.01 & 0.990 $\pm$ 0.00 & +0.031 & 0.581 $\pm$ 0.01 & 0.713 $\pm$ 0.01 & +0.132 \\
\midrule
\multirow{2}{*}{Native-thinking}
& DeepSeek-R1-Distill-7B & 0.749 $\pm$ 0.02 & 1.000 $\pm$ 0.00 & +0.251 & 0.546 $\pm$ 0.03 & 0.790 $\pm$ 0.03 & +0.244 \\
& Qwen3-30B-A3B-Thinking & 0.956 $\pm$ 0.02 & 1.000 $\pm$ 0.00 & +0.044 & 0.855 $\pm$ 0.02 & 0.632 $\pm$ 0.01 & $-$0.223 \\
\bottomrule
\end{tabular}
\caption{Base-to-tuned changes. Native-thinking deltas use a larger generation budget.}
\label{tab:gains}
\end{table*}

Fine-tuning reliably eliminates formatting errors: all adapted open-weight checkpoints achieve 99.0\%--100.0\% FFR across both precise and categorical splits, accompanied by perfect decline accuracy (100.0\%) on unanswerable questions. For standard instruct models, this formatting alignment translates into substantial end-to-end improvements: non-thinking checkpoints gain between +0.132 and +0.311 in E2E (Table~\ref{tab:gains}), with Gemma-4-31B-IT reaching an E2E of 0.867.
ChatTS records an E2E of 0.876; however, because ChatTS was evaluated via a single deterministic run rather than our three-seed evaluation protocol, its score should be viewed as competitive with dense language models rather than statistically superior.

\paragraph{Architecture and Reasoning Dynamics.}
Our results highlight two unexpected patterns:
\begin{enumerate}[leftmargin=*]
  \item \textbf{Scale does not dictate task mastery:} Adapted Qwen3-8B (0.816 E2E) matches the $4\times$ larger Qwen3-32B (0.825 E2E). Moreover, evaluated MoE checkpoints (Qwen3-30B-A3B) trail dense models of comparable total scale, reaching only 0.632--0.713 E2E. As post-training mixtures and active parameter footprints (3B active) differ, we report this as an empirical observation rather than a causal architectural claim.
  \item \textbf{Direct SFT can impair native reasoning models:} While DeepSeek-R1-Distill-7B gains +0.244 E2E post-SFT, Qwen3-30B-A3B-Thinking drops 0.223 E2E ($0.855 \to 0.632$). As discussed in Section~\ref{sec:discussion}, fine-tuning reasoning models on direct-answer pairs inadvertently suppresses internal scratchpad generation.
\end{enumerate}

\subsection{Fraud QA: Verdict--Evidence Separation}
\label{sec:fraudqa-results}

Table~\ref{tab:fraudqa-proprietary} reports Fraud QA performance for six proprietary models, each run with thinking on and off (12 configurations, three seeds each; all produce parseable structured output, FFR $\geq 99.3\%$). Verdict accuracy is consistently high---averaging $73.1\%$ and peaking at $87.7\%$ (GPT-5 mini, thinking off)---yet \emph{joint} accuracy, which additionally requires the cited evidence to exactly match the gold rationale, averages only $12.5\%$ and never exceeds $22.8\%$. Evidence grounding, not verdict calling, is the bottleneck: even on the rows where evidence is wrong, citation precision never exceeds $34.7\%$ and recall tops out at $85.6\%$. Because the task is label-balanced ($50\%$ fraud), configurations such as Claude Sonnet and Haiku (thinking on, $53.5$--$56.2\%$ verdict accuracy) sit only modestly above the $50\%$ always-fraud floor, reflecting a hard-negative wall in which benign look-alike cases are over-flagged. Thinking mode trades verdict accuracy for evidence-grounded correctness in every matched pair: it raises joint accuracy in all six model pairs while lowering verdict accuracy in all six (e.g., GPT-5 mini: joint $1.0\to15.4$, verdict $87.7\to82.5$; Gemini 3.1 Flash-Lite: joint $14.1\to22.8$, verdict $79.7\to69.0$). Finally, nominal model tier is not monotonic with performance: the smallest-tier model, Gemini 3.1 Flash-Lite (thinking), attains the best joint accuracy ($22.8\%$), ahead of every flagship configuration.

\begin{table*}[t]
\centering
\small
\renewcommand{\arraystretch}{1.06}
\setlength{\tabcolsep}{3.5pt}
\begin{tabular}{lcccccc}
\toprule
\multirow{2}{*}{\textbf{Model}} & \multicolumn{3}{c}{\textbf{Precise QA (\%)}} & \multicolumn{3}{c}{\textbf{Evidence Quality}} \\
\cmidrule(lr){2-4} \cmidrule(lr){5-7}
 & \textbf{FFR} & \textbf{Verdict Acc.} & \textbf{Joint Acc.} & \textbf{Rec.\ (\%)} & \textbf{Prec.\ (\%)} & \textbf{Over-citation} \\
\midrule
GPT-5 (thinking on) & $\mathbf{100.0 \pm 0.0}$ & $80.9 \pm 1.2$ & $22.4 \pm 0.6$ & $58.2 \pm 0.9$ & $34.0 \pm 0.6$ & $1.77 \pm 0.05$ \\
GPT-5 (thinking off) & $\mathbf{100.0 \pm 0.0}$ & $81.4 \pm 0.4$ & $0.9 \pm 0.3$ & $74.2 \pm 1.4$ & $34.5 \pm 0.4$ & $2.38 \pm 0.01$ \\
GPT-5 mini (thinking on) & $\mathbf{100.0 \pm 0.0}$ & $82.5 \pm 0.6$ & $15.4 \pm 0.2$ & $66.3 \pm 1.2$ & $\mathbf{34.7 \pm 0.2}$ & $1.83 \pm 0.04$ \\
GPT-5 mini (thinking off) & $\mathbf{100.0 \pm 0.0}$ & $\mathbf{87.7 \pm 1.0}$ & $1.0 \pm 0.4$ & $\mathbf{85.6 \pm 0.2}$ & $34.4 \pm 0.3$ & $2.75 \pm 0.01$ \\
Claude Opus 4.6 (thinking on) & $\mathbf{100.0 \pm 0.0}$ & $71.7 \pm 0.3$ & $16.5 \pm 0.2$ & $62.8 \pm 0.4$ & $28.8 \pm 0.3$ & $2.28 \pm 0.01$ \\
Claude Opus 4.6 (thinking off) & $\mathbf{100.0 \pm 0.0}$ & $79.8 \pm 0.2$ & $8.3 \pm 0.4$ & $64.3 \pm 0.7$ & $33.1 \pm 0.4$ & $2.13 \pm 0.00$ \\
Claude Sonnet 4.6 (thinking on) & $\mathbf{100.0 \pm 0.0}$ & $53.5 \pm 0.7$ & $10.5 \pm 0.6$ & $59.0 \pm 1.1$ & $30.8 \pm 0.4$ & $1.95 \pm 0.00$ \\
Claude Sonnet 4.6 (thinking off) & $99.3 \pm 0.4$ & $64.8 \pm 1.0$ & $9.1 \pm 0.2$ & $61.0 \pm 0.8$ & $31.6 \pm 0.2$ & $2.08 \pm 0.01$ \\
Claude Haiku 4.5 (thinking on) & $\mathbf{100.0 \pm 0.0}$ & $56.2 \pm 0.9$ & $18.5 \pm 0.4$ & $43.9 \pm 2.1$ & $27.3 \pm 0.3$ & $1.66 \pm 0.03$ \\
Claude Haiku 4.5 (thinking off) & $\mathbf{100.0 \pm 0.0}$ & $69.7 \pm 0.6$ & $10.5 \pm 0.3$ & $50.3 \pm 0.3$ & $32.2 \pm 0.4$ & $1.64 \pm 0.02$ \\
Gemini 3.1 Flash-Lite (thinking on) & $\mathbf{100.0 \pm 0.0}$ & $69.0 \pm 1.7$ & $\mathbf{22.8 \pm 0.5}$ & $39.8 \pm 0.6$ & $27.9 \pm 0.4$ & $1.55 \pm 0.07$ \\
Gemini 3.1 Flash-Lite (thinking off) & $\mathbf{100.0 \pm 0.0}$ & $79.7 \pm 0.5$ & $14.1 \pm 0.4$ & $27.8 \pm 0.8$ & $30.4 \pm 1.6$ & $\mathbf{1.17 \pm 0.06}$ \\
\bottomrule
\end{tabular}
\caption{FraudQA performance of proprietary models (zero-shot), mean $\pm$ SD over 3 seeds. Left: answer-level accuracy over all rows (\textbf{Precise QA}). Right: \textbf{Evidence Quality} conditioned on the evidence-incorrect population ($P \neq G$). Bold indicates best per column.}
\label{tab:fraudqa-proprietary}
\end{table*}

\paragraph{Case Study: Why the Verdict--Evidence Gap?}
The gap is largest for GPT-5 mini (thinking off), which attains the highest verdict accuracy in our suite ($87.7\%$) yet almost never cites exactly the gold evidence (joint accuracy $1.0\%$)---an $86.7$-point gap.
The cause is not missing evidence but \emph{over-citation}.
On confirmed-fraud cases it labels correctly, it recovers the full gold set $63\%$ of the time and attains the highest evidence recall of any configuration ($0.86$); yet it cites $7.8$ features on average where the gold rationale needs only two to five, over-citing at $2.8\times|G|$ and dragging precision to $0.34$.
On one self-dealing case (gold: \texttt{seq\_sndr\_email}, \texttt{rcvr\_email}, \texttt{seq\_amount}, \texttt{seq\_um\_txn\_d30}), it lists eleven features, adding merchant metadata (\texttt{merch\_country}, \texttt{merchant\_category}, \texttt{merch\_total\_txns}, \dots) unrelated to the self-payment pattern.
The model thus identifies decisive signals but hedges by enumerating every plausibly-relevant field rather than isolating the minimal causal set. Chain-of-thought reasoning is a partial corrective---it prunes over-citation ($2.8\to1.8\times$) and raises joint accuracy ($1.0\to15.4\%$) at some cost to recall---indicating the gap reflects a selection/decisiveness failure rather than a knowledge gap.

\subsection{Answering by Querying the Records}
\label{sec:sql-baseline}

We also test a tool-use protocol in which the model never sees the transaction text. It receives the schema of a per-merchant SQLite database that holds exactly what the text shows: the merchant fields, plus one row per transaction with its displayed date, hour, weekday, and amount. The model writes one query and sees the result, with one retry if the query fails. It then writes the answer in the required format. We compare this protocol with the direct protocol of Section~\ref{sec:protocol} on six open-weight models.

This comparison uses an expanded question set with 71 types: the 27 original types and 44 new ones, such as year-over-year changes and rankings of weekdays, months, hours, and amounts. Every gold answer is computed from the rendered text the model reads. The balanced test set has 426 questions over 408 merchants, sampled the same way as original test items and from the same pool of held-out test merchants. For 59 of these questions the gold answer is \textsc{unknown}. Because the question set differs, we report these results separately from the original evaluation (Table~\ref{tab:sql-main}).

\begin{table}[t]
\centering
\small
\setlength{\tabcolsep}{8.0pt}
\begin{tabular}{lccc}
\toprule
Model & SQL & Direct & Direct, 32k \\
\midrule
\multicolumn{4}{l}{\emph{Answering without reasoning}} \\
Qwen3-8B & 78.7 & 50.6 & -- \\
Qwen3-32B & 76.7 & 55.6 & -- \\
Qwen3-30B-A3B-Instruct & 87.6 & 59.1 & -- \\
\midrule
\multicolumn{4}{l}{\emph{Answering with reasoning}} \\
Qwen3-30B-A3B-Thinking & 87.9 & 77.4 & 84.9 \\
GPT-OSS-20B & 86.3 & 82.4 & \textbf{85.5} \\
GPT-OSS-120B & \textbf{90.5} & \textbf{83.3} & 83.9 \\
\bottomrule
\end{tabular}
\caption{End-to-end accuracy (E2E, \%) on the expanded 71-type test set (426 questions). SQL: the model queries a database containing the same records. Direct: protocol with default 8,192-token budget, or 32,000 tokens for reasoning models. Values are means over 3 seeds.}
\label{tab:sql-main}
\end{table}

SQL access helps every model. The three models that answer without reasoning gain 21--28 points of E2E, and Qwen3-30B-A3B-Instruct rises from 59.1 to 87.6, matching its reasoning counterpart under SQL (87.9). The reasoning models gain less at the default budget: 10.5 points for Qwen3-30B-A3B-Thinking, 3.9 for GPT-OSS-20B, and 7.2 for GPT-OSS-120B. Part of their deficit comes from truncated reasoning. In seed 0, 46 of Qwen3-30B-A3B-Thinking's 426 direct answers run out of tokens before the answer. With 32,000 tokens, its gap to SQL falls to 3.0 points and GPT-OSS-20B's to 0.8. GPT-OSS-120B rarely ran out (3 of 426 in seed 0) and keeps a gap of 6.5. With SQL, it is the strongest configuration (90.5).

For models that answer without reasoning, much of the direct-protocol error comes from computing over rendered records, and exact database access removes most of it. Two caveats apply: (1) the SQL prompt adds two notes about SQLite behavior (without them, Qwen3-8B scores 75.9 instead of 78.7); (2) SQL access changes how the model reaches the records as well as their format. Appendix~\ref{app:sql} gives full protocol details and all metrics.

\section{Discussion}
\label{sec:discussion}

The core contribution of \benchmark lies in the intersection of four challenges: irregular temporal event data, dynamic query-dependent logic, strictly typed output, and feature-level risk attribution (Section~\ref{sec:related-work}). While existing benchmarks address subsets of these properties, transaction analysis requires models to solve them simultaneously. In this context, separating performance into format compliance (FFR), conditional accuracy (CA), and end-to-end correctness (E2E) provides actionable operational diagnostics.

Our empirical findings demonstrate that raw parameter scale does not reliably predict performance on irregular event sequences (Section~\ref{sec:results}). Smaller, fine-tuned dense models such as Qwen3-8B frequently match or outperform checkpoints with four to thirty times more parameters. While architectural differences, training mixtures, and hyperparameter choices prevent broad causal claims regarding dense versus sparse models, these results hold practical value: targeted domain adaptation provides a viable, computationally efficient path toward operational transaction QA without mandating massive parameter footprints.

Our empirical results also provide a cautionary note against viewing supervised fine-tuning as a universal remedy for multi-step reasoning failures. Across identical non-thinking configurations, LoRA adapters improve E2E almost always, driving FFR near 100\%, but leaving residual errors mostly on arithmetic and relational calculations. Furthermore, fine-tuning can actively harm models that rely on extended test-time compute: Qwen3-30B-A3B-Thinking suffered a 22.3-point decline in E2E following adaptation. Because this checkpoint was fine-tuned on direct-answer pairs rather than reasoning traces, the supervision signal appears to suppress its internal scratchpad generation.

The fraud results carry a direct operational implication: a correct verdict is not evidence of grounded reasoning. Against gold evidence carefully defined and verified for each scenario, verdict accuracy nears $88\%$ while strict \emph{joint} accuracy never exceeds $22.8\%$, and citation precision stays below $35\%$ on missed rows. Models tend to reach the right conclusion without selecting the right supporting evidence. We interpret this as a measure of evidence-\emph{selection} quality: because the gold rationale is specified as a set of discriminative feature names rather than a full natural-language reasoning trace, selecting them is necessary but not sufficient for a faithful explanation. The gap we report is therefore a conservative lower bound on the grounding problem.

\section{Conclusion}
\label{sec:conclusion}

We introduced \benchmark, a benchmark for single-turn analytical QA over irregular merchant transaction histories with a controlled fraud-risk feature-selection stress test. Across six LoRA-adapted open-weight checkpoints, E2E point estimates range from 0.632 to 0.867, with Gemma-4-31B-IT highest among them. Decomposed metrics show adaptation often improves format compliance without eliminating answering inaccuracies, and that high fraud-verdict accuracy can coexist with weak supporting-feature selection. These findings motivate evaluation that separates computation, rendering, and evidence.

\section*{Limitations}
\label{sec:limitations}

The benchmark is synthetic data generated by fitting REaLTabFormer to approved internal merchant-transaction data; marginal and pairwise fidelity diagnostics (Section~\ref{sec:construction}) do not establish downstream behavioral equivalence, fairness, or transfer to new markets, currencies, or payment rails, and REaLTabFormer is not a differentially private mechanism---exact-copy and distance-to-closest-record checks are empirical diagnostics, not formal privacy guarantees.

The evaluated populations are small (162 general-QA and 500 fraud-QA items; 426 questions for the SQL comparison of Section~\ref{sec:sql-baseline}), and reported standard deviations reflect variability across three decoding seeds, not uncertainty over test items. Reported E2E values are an approximation ($\mu_{FFR}\times\mu_{CA}$ across seed-averaged means), not the harness-computed per-item quantity, except in Tables~\ref{tab:sql-main} and~\ref{tab:sql-full}.

Training and test questions are drawn from the same 27-task-family, 3-wording-variant template distribution and the same synthetic generator; held-out or independently paraphrased template evaluation is an open direction.

Fraud QA covers five scenario families curated by fraud-prevention experts and abstracted from real cases into canonical patterns. It is deliberately scoped to test an LLM's basic ability to recognize fraud patterns rather than to resolve full operational complexity. The task is conducted in English and in a single turn.

Model comparisons are descriptive, not causal: model families, post-training recipes, adapter footprints, hyperparameters, output-token budgets, tokenizers, and serving backends differ across checkpoints.

\section*{Ethical Considerations}

The underlying real transaction data used to fit the synthetic generator is sensitive and was accessed under internal approval; it is not released, and only the synthetic, privacy-mitigated derivative described above is intended for release. The controlled fraud-risk scenarios are rule-defined and authored for this benchmark; they must not be interpreted as validated real-world fraud labels, and models evaluated on this benchmark must not be deployed as autonomous fraud, credit, compliance, or adverse-action systems. This benchmark is intended for research use in evaluating language-conditioned analytical and evidence-selection capabilities, with human oversight assumed for any higher-stakes downstream use.

\bibliographystyle{plainnat}
\bibliography{ref}

\clearpage
\appendix

\section{Analytical Question Types and Format Variants}
\label{app:questions}

We define 27 analytical question types, each with three wording variants (Table~\ref{tab:generalqa}) and 43 answer-format variants across precise and categorical regimes (Table~\ref{tab:format_variants}). Three question types (9, 24, 25) are not answerable from the transaction text alone (no user-level information); their gold answer is \textsc{unknown}, scored via decline accuracy.

\paragraph{Fraud scenarios.}
\emph{Story 1 (Merchant Registration Linkage):} a synthetic-looking email domain plus a large same-day cluster of merchant registrations under that domain may indicate coordinated batch registration; the matched negative preserves the registration count but uses a widely used public email provider.
\emph{Story 2 (Sender--Receiver Identity Linkage):} a sender email matching the merchant's receiver email, combined with unusually high transaction velocity/value from that sender, may indicate a collusive relationship; the matched negative preserves comparable activity but removes the email match. Reference verdicts and evidence features were supplied by a domain expert at the source payment platform.

\paragraph{Transaction features.}
Merchant-level: category, country, account age, total transaction count, distinct buyers (plus, for fraud QA: registration date, receiver email, same-domain same-day registration count, email domain). Per-transaction: datetime, hour, amount (plus, for fraud QA: buyer ID, sender email, user--merchant transaction count/spend in the past 7/30 days).

{\small
\begin{longtable}{p{2.6cm}p{3.8cm}p{3.8cm}p{3.8cm}}
\caption{27 analytical question types, 3 wording variants each. $^\dagger$Not answerable from rendered text alone; gold label is \textsc{unknown}.} \label{tab:generalqa} \\
\toprule
\textbf{Type} & \textbf{Variant 1} & \textbf{Variant 2} & \textbf{Variant 3} \\
\midrule
\endfirsthead
\multicolumn{4}{c}{\tablename\ \thetable\ -- \textit{Continued from previous page}} \\
\toprule
\textbf{Type} & \textbf{Variant 1} & \textbf{Variant 2} & \textbf{Variant 3} \\
\midrule
\endhead
\midrule
\multicolumn{4}{r}{\textit{Continued on next page}} \\
\endfoot
\bottomrule
\endlastfoot
1. merchant category & What category does this merchant operate in? & How would you classify this merchant's business? & What kind of goods or services does this merchant sell?\\
2. merchant country & Where is this merchant based? & What country does this merchant operate from? & In which country is this merchant located?\\
3. busiest day & On which weekday is this merchant most active? & What's the peak day of the week for transactions? & Which day sees the highest transaction volume?\\
4. busiest time of day & When during the day is this merchant busiest? & Which time of day sees the most activity? & What part of the day has peak volume?\\
5. merchant age & How old is this merchant's account? & How long has this merchant been operating? & When was this merchant first established?\\
6. transaction count & What's this merchant's total transaction count? & How many payments has this merchant processed? & What's the overall transaction volume?\\
7. distinct buyers & How many unique customers has this merchant served? & What's the number of different buyers? & How many separate buyers transacted here?\\
8. txns per buyer & What's the average number of transactions per buyer? & How many times does a typical buyer transact? & On average, how many purchases does each customer make?\\
9. buyer concentration$^\dagger$ & How concentrated are transactions among the top buyer? & What fraction of activity comes from the busiest single buyer? & What percentage of transactions does the top buyer account for?\\
10. typical amount & What's the average transaction value? & How much is a typical purchase here? & What's the mean amount per transaction?\\
11. minimum amount & What's the lowest-value transaction? & How small is the smallest payment? & What's the minimum amount transacted?\\
12. maximum amount & What's the highest-value transaction? & How large is the biggest payment? & What's the maximum amount transacted?\\
13. largest vs average & How many times bigger is the largest transaction than the average? & What's the ratio of the biggest transaction to the typical one? & By what factor does the largest transaction exceed the average?\\
14. amount CoV & How spread out are the transaction amounts relative to the mean? & How consistent are this merchant's transaction values? & What's the relative variability of the amounts?\\
15. median amount & What's the middle (50th-percentile) transaction value? & Half of transactions fall below what amount? & What's the median spend per transaction?\\
16. transaction frequency & What's the typical gap between transactions? & How frequently does this merchant transact? & On average, how much time passes between payments?\\
17. interval burstiness & Are transactions evenly spaced or clumped together? & How consistent is the timing between transactions? & How variable are the gaps between transactions?\\
18. longest inactivity gap & What's the longest stretch with no transactions? & How long was the biggest pause in activity? & What's the maximum time between consecutive transactions?\\
19. windowed count & What's the transaction count over the past 30 days? & How many payments were made in the last month? & How active has this merchant been in the last 30 days?\\
20. busiest hour & At what hour of the day are transactions most frequent? & What's the peak hour for this merchant? & Which hour sees the highest transaction volume?\\
21. most recent transaction & When did this merchant last transact? & What's the date of the latest transaction? & How recently did this merchant transact?\\
22. first transaction & When did this merchant's activity begin? & What's the date of the earliest transaction? & When did this merchant make its first sale?\\
23. transacting days & On how many separate days was this merchant active? & How many distinct days saw at least one transaction? & On how many different days did this merchant transact?\\
24. buyer account age$^\dagger$ & What's the typical account age of this merchant's buyers? & How established are this merchant's customers? & How old are the buyers' accounts, on average?\\
25. buyer countries$^\dagger$ & From how many different countries do buyers originate? & How geographically spread are this merchant's buyers? & What's the number of distinct buyer countries?\\
26. amount trend & Are transaction amounts rising or falling over time? & Is the spend per transaction growing or shrinking? & How has the typical amount changed over the history?\\
27. activity trend & Is this merchant's transaction volume rising or falling over time? & Is activity picking up or slowing down? & How has the transaction count changed over time?\\
\end{longtable}
}

\begin{table}[t]
\centering
\small
\begin{tabular}{p{2.0cm}p{1.8cm}p{8.5cm}}
\toprule
\textbf{Type(s)} & \textbf{Regime} & \textbf{Example format instruction} \\
\midrule
1--4, 20--22 & categorical & Category / country / day-code / part-of-day / clock-hour / ISO or prose date \\
5, 6--9, 13--19, 23, 26--27 & precise & Exact count, ratio, coefficient of variation, or interval, per type \\
10--12, 15 & precise & Exact dollar amount, full digits or K/M-suffixed \\
24, 25 & precise & Exact age (years/months/days) or country count \\
\bottomrule
\end{tabular}
\caption{43 answer-format variants across the 27 question types (condensed), spanning precise and categorical regimes.}
\label{tab:format_variants}
\end{table}

\section{Full Metric Definitions}
\label{app:metrics}

Realism (numerical): $D=\max_x|F_{\mathrm{real}}(x)-F_{\mathrm{syn}}(x)|$ (Kolmogorov--Smirnov). Realism (categorical): $\mathrm{TVD}=\tfrac{1}{2}\sum_i|P(x_i)-Q(x_i)|$. Fraud verdict accuracy, precision, and recall, and macro-averaged evidence precision/recall, with $E_{\mathrm{pred}}^{(i)}, E_{\mathrm{gold}}^{(i)}$ the predicted/gold evidence-feature sets for case $i$:
\begin{align*}
\mathrm{Acc} &= \tfrac{1}{N}\textstyle\sum_i \mathbf{1}[C_{\mathrm{pred}}^{(i)}=C_{\mathrm{gold}}^{(i)}] \\
P_{\mathrm{verdict}} &= \tfrac{\sum_i \mathbf{1}[C_{\mathrm{pred}}^{(i)}=1 \wedge C_{\mathrm{gold}}^{(i)}=1]}{\sum_i \mathbf{1}[C_{\mathrm{pred}}^{(i)}=1]} \\
R_{\mathrm{verdict}} &= \tfrac{\sum_i \mathbf{1}[C_{\mathrm{pred}}^{(i)}=1 \wedge C_{\mathrm{gold}}^{(i)}=1]}{\sum_i \mathbf{1}[C_{\mathrm{gold}}^{(i)}=1]} \\
P_{\mathrm{evid}} &= \tfrac{1}{N}\textstyle\sum_i \tfrac{|E_{\mathrm{pred}}^{(i)}\cap E_{\mathrm{gold}}^{(i)}|}{|E_{\mathrm{pred}}^{(i)}|}, \;\;
R_{\mathrm{evid}} = \tfrac{1}{N}\textstyle\sum_i \tfrac{|E_{\mathrm{pred}}^{(i)}\cap E_{\mathrm{gold}}^{(i)}|}{|E_{\mathrm{gold}}^{(i)}|}
\end{align*}
As noted in Section~\ref{sec:protocol}, these are undefined when $|E_{\mathrm{pred}}^{(i)}|=0$ or $|E_{\mathrm{gold}}^{(i)}|=0$ respectively.

\section{Fine-Tuning Configuration}
\label{app:hyperparams}

\begin{table}[h!]
\centering
\small
\begin{tabular}{lrrrrr}
\toprule
Checkpoint & Params & Type & Epochs & LR & Trainable \\
\midrule
DeepSeek-R1-Distill-Qwen-7B & 7.62B & Dense & 3 & $1{\times}10^{-4}$ & 80.7M \\
Qwen3-8B & 8B & Dense & 3 & $1{\times}10^{-4}$ & 87.3M \\
Qwen3-32B & 32B & Dense & 2 & $7{\times}10^{-5}$ & 268M \\
Qwen3-30B-A3B-Instruct & 30B/3B & MoE & 2 & $7{\times}10^{-5}$ & 1.99B \\
Qwen3-30B-A3B-Thinking & 30B/3B & MoE & 2 & $7{\times}10^{-5}$ & 1.99B \\
Gemma-4-31B-IT & 31B & Dense & 2 & $5{\times}10^{-5}$ & 244M \\
\bottomrule
\end{tabular}
\caption{LoRA configurations (rank 32, $\alpha=64$, dropout 0.00--0.05).}
\label{tab:models}
\end{table}

All checkpoints used AdamW ($\beta_1{=}0.9,\beta_2{=}0.999,\epsilon{=}10^{-8}$), zero weight decay, gradient clipping at norm 1.0, cosine LR decay with 0.03 warmup ratio, bf16 precision, a global seed of 0, final-epoch checkpoint selection, and a 4096-token max sequence length with left-truncation (text models) or example-dropping (ChatTS). Evaluation used greedy decoding (temperature 0) for the SFT harness's own internal checks and temperature 0.7 for the 3-seed reported results (Section~\ref{sec:protocol}). Hardware: 8$\times$NVIDIA B200 (183GB) per node, DDP via \texttt{torchrun}.

\section{Full Zero-Shot and Fraud QA Results (Open-Weight Models)}
\label{app:full_results}

\begin{table*}[t]
\centering
\small
\setlength{\tabcolsep}{6.0pt}
\begin{tabular}{llccccc}
\toprule
Open-weight checkpoint & Thinking & \multicolumn{3}{c}{Precise QA} & \multicolumn{2}{c}{Categorical QA} \\
\cmidrule(lr){3-5}\cmidrule(lr){6-7}
 & & FFR & CA (strict) & E2E (approx.) & FFR & CA (match) \\
\midrule
Qwen3-8B & Off & \textbf{98.1} & 51.5 & 50.5 & 88.9 & 66.7 \\
Qwen3-8B & On  & 93.7 & 87.1 & 81.6 & 95.4 & 66.0 \\
Qwen3-32B & Off & 97.8 & 56.8 & 55.5 & 99.1 & 68.2 \\
Qwen3-32B & On  & 94.0 & 84.8 & 79.7 & \textbf{100.0} & 63.0 \\
Qwen3-30B-A3B-Instruct-2507 & Off & 95.9 & 60.6 & 58.1 & \textbf{100.0} & 63.0 \\
Qwen3-30B-A3B-Thinking-2507 & On  & 95.6 & 89.4 & \textbf{85.5} & 97.2 & 72.4 \\
DeepSeek-R1-Distill-7B & On$^\dagger$ & 74.9 & 72.9 & 54.6 & 87.0 & 58.6 \\
Gemma-4-31B-IT & Off & 95.6 & 74.4 & 71.1 & \textbf{100.0} & 66.7 \\
Gemma-4-31B-IT & On  & 94.9 & 87.0 & 82.6 & 97.2 & 75.2 \\
GPT-OSS-20B & Off & 84.1 & 84.2 & 70.8 & 99.1 & 69.2 \\
GPT-OSS-120B & Off & 81.9 & 87.2 & 71.4 & 95.4 & 73.8 \\
GPT-OSS-120B & On  & 82.9 & 88.9 & 73.7 & 88.9 & 70.8 \\
Qwen3-235B & Off & 95.2 & 62.7 & 59.7 & 83.3 & \textbf{76.7} \\
Qwen3-235B & On  & 93.7 & 87.1 & 81.6 & \textbf{100.0} & 72.2 \\
GLM-5.2 & Off & 94.6 & 72.5 & 68.6 & \textbf{100.0} & 68.5 \\
GLM-5.2 & On  & 92.4 & 88.0 & 81.3 & 98.1 & 67.9 \\
DeepSeek-V4-Flash & Off & 73.7 & 92.2 & 68.0 & 92.6 & 71.0 \\
DeepSeek-V4-Flash & On  & 92.4 & 87.3 & 80.7 & 92.6 & 73.0 \\
DeepSeek-V4-Pro & Off & 73.0 & \textbf{93.9} & 68.5 & 94.4 & 70.6 \\
DeepSeek-V4-Pro & On  & 93.3 & 90.8 & 84.7 & 95.4 & 70.9 \\
Kimi-K2.6 & Off & 64.1 & 92.2 & 59.1 & 82.4 & 71.9 \\
Kimi-K2.6 & On  & 88.9 & 87.5 & 77.8 & 93.5 & 74.3 \\
\bottomrule
\end{tabular}
\caption{Zero-shot open-weight performance. $^\dagger$Reasoning-only checkpoint.}
\label{tab:opensource-zeroshot}
\end{table*}

\section{SQL Baseline Details}
\label{app:sql}

\paragraph{Database.}
Each merchant's database has a one-row \texttt{merchant} table (category, country, account age, total transactions, distinct buyers) and a \texttt{transactions} table with one row per rendered transaction (index, displayed date and time, date, hour, weekday, amount, and a timestamp for ordering). The schema description states that the records end at the latest transaction, which serves as ``now'', and that buyer identities are not available. It also defines chronological order by timestamp. The direct protocol states none of this; its model infers it from listed transactions. SQLite is extended with \texttt{median()} and \texttt{stddev()}, as analytics databases provide. Queries run read-only with a 5-second timeout, and the model sees at most 20 result rows. Hand-written reference queries reproduce the gold answer for 18 question types on every validation merchant, confirming database sufficiency.

\paragraph{Generation.}
Each question gets one sample per seed (seeds 0--2, temperature 0.7), with one retry after a failed query and no voting over samples. Qwen3-8B and Qwen3-32B run without thinking in both protocols, and Qwen3-30B-A3B-Instruct has no reasoning mode. These models receive 512 tokens per query and 128 for the answer. Qwen3-30B-A3B-Thinking and GPT-OSS always reason and receive 8,192 tokens for each query attempt and for the answer. GPT-OSS runs at its chat template's default reasoning effort (medium), and vLLM's Marlin kernel serves its 4-bit (MXFP4) expert layers. All runs used vLLM 0.19.0 on a single NVIDIA B200.

\paragraph{Tool-behavior notes.}
In early Qwen3-8B traces, the model wrote PostgreSQL median syntax, which SQLite rejects, and repeated it after the error. It also filled answer templates with ``+/-0\%'' when a query returned NULL. The main setup therefore adds two notes that describe the tool. After a failed query, a reminder states that the database is SQLite and provides \texttt{median()} and \texttt{stddev()}. In the answer step, a statement notes that a NULL or empty result indicates records do not contain the value. With the notes, all 12 median questions in Qwen3-8B's seed 0 obtain a working query; without them, none do (Table~\ref{tab:sql-full}, ``SQL, no notes''). The NULL note has one known cost: \texttt{SUM()} returns NULL when no rows match, so on the 2 test questions whose date range contains no sales (gold \$0), models following the note answer \textsc{unknown}, costing approximately 0.5 E2E points.

\paragraph{Reasoning budget.}
With 32,000 tokens, 44 of Qwen3-30B-A3B-Thinking's 46 truncated direct answers (seed 0) finish, and the 380 answers that finished within 8,192 tokens emerge identical. Its gain therefore comes entirely from newly finished answers. Exact-value accuracy falls from 93.9 to 89.5 as E2E rises. GPT-OSS behaves differently: it does not reproduce answers token for token when only the budget changes. In seed 0, 265 of 415 previously finished answers are identical for GPT-OSS-20B, and 290 of 423 for GPT-OSS-120B, mixing newly finished answers (11 and 3) with run-to-run variation.

\begin{table*}[t]
\centering
\small
\setlength{\tabcolsep}{3.2pt}
\begin{tabular}{llccccc}
\toprule
Model & Method & E2E & FFR & CA exact (strict) & CA categorical & Decline acc. \\
\midrule
Qwen3-8B & SQL & 78.7 $\pm$ 1.1 & 92.2 $\pm$ 1.3 & 84.6 $\pm$ 2.2 & 83.8 $\pm$ 3.1 & 44.1 $\pm$ 3.4 \\
 & SQL, no notes & 75.9 $\pm$ 1.9 & 88.8 $\pm$ 2.0 & 84.9 $\pm$ 1.8 & 83.4 $\pm$ 3.4 & 35.0 $\pm$ 8.4 \\
 & Direct & 50.6 $\pm$ 0.2 & 96.9 $\pm$ 0.4 & 46.4 $\pm$ 0.1 & 61.2 $\pm$ 1.0 & 18.1 $\pm$ 1.0 \\
\midrule
Qwen3-32B & SQL & 76.7 $\pm$ 2.0 & 88.7 $\pm$ 0.3 & 86.2 $\pm$ 2.6 & 84.3 $\pm$ 0.6 & 61.6 $\pm$ 7.6 \\
 & Direct & 55.6 $\pm$ 0.3 & 97.6 $\pm$ 0.4 & 49.4 $\pm$ 0.5 & 69.9 $\pm$ 1.1 & 51.4 $\pm$ 5.4 \\
\midrule
Qwen3-30B-A3B-Instruct & SQL & 87.6 $\pm$ 0.4 & 95.5 $\pm$ 1.1 & 91.9 $\pm$ 0.8 & 89.6 $\pm$ 0.6 & 59.9 $\pm$ 2.0 \\
 & Direct & 59.1 $\pm$ 0.7 & 99.1 $\pm$ 0.3 & 50.9 $\pm$ 0.4 & 72.1 $\pm$ 1.2 & 42.9 $\pm$ 2.0 \\
\midrule
Qwen3-30B-A3B-Thinking & SQL & 87.9 $\pm$ 0.8 & 94.4 $\pm$ 0.9 & 93.6 $\pm$ 0.5 & 91.2 $\pm$ 0.8 & 81.4 $\pm$ 4.5 \\
 & Direct & 77.4 $\pm$ 1.4 & 83.6 $\pm$ 1.8 & 93.9 $\pm$ 0.8 & 88.0 $\pm$ 1.6 & 84.2 $\pm$ 1.0 \\
 & Direct, 32k & 84.9 $\pm$ 0.6 & 95.1 $\pm$ 0.8 & 89.5 $\pm$ 0.5 & 86.8 $\pm$ 1.1 & 84.2 $\pm$ 1.0 \\
\midrule
GPT-OSS-20B & SQL & 86.3 $\pm$ 0.4 & 92.3 $\pm$ 0.4 & 94.0 $\pm$ 0.9 & 91.8 $\pm$ 1.3 & 86.4 $\pm$ 4.5 \\
 & Direct & 82.4 $\pm$ 0.3 & 92.4 $\pm$ 0.5 & 89.5 $\pm$ 0.6 & 86.2 $\pm$ 3.2 & 79.1 $\pm$ 2.0 \\
 & Direct, 32k & 85.5 $\pm$ 0.4 & 95.8 $\pm$ 0.4 & 89.4 $\pm$ 1.5 & 86.0 $\pm$ 3.1 & 79.7 $\pm$ 0.0 \\
\midrule
GPT-OSS-120B & SQL & 90.5 $\pm$ 0.3 & 97.1 $\pm$ 0.3 & 94.9 $\pm$ 0.4 & 87.8 $\pm$ 2.4 & 82.5 $\pm$ 2.6 \\
 & Direct & 83.3 $\pm$ 0.4 & 91.7 $\pm$ 0.8 & 90.8 $\pm$ 0.6 & 89.5 $\pm$ 1.3 & 81.9 $\pm$ 1.0 \\
 & Direct, 32k & 83.9 $\pm$ 1.5 & 92.3 $\pm$ 0.3 & 91.8 $\pm$ 1.4 & 88.1 $\pm$ 2.7 & 81.9 $\pm$ 1.0 \\
\bottomrule
\end{tabular}
\caption{SQL baseline and direct answers on the expanded 71-type test set, all metrics (\%, mean $\pm$ SD over 3 seeds). E2E and FFR are over 367 answerable questions; CA is over format-valid answers; decline accuracy is over 59 questions whose gold is \textsc{unknown}.}
\label{tab:sql-full}
\end{table*}

\paragraph{Worked Example.}
To illustrate how structured transaction records are serialized and queried in \benchmark, consider the following sample instance (hand-constructed for exposition):
\begin{quote}
\small
\textbf{Serialized Context (truncated):} \textit{``This merchant operates in the restaurants category and is based in the United States, with an account age of approximately 2.3 years. On Monday, March 3, 2025 at 12:15pm, a transaction of \$24.50 was processed. On Tuesday, March 4, 2025 at 7:40pm, a transaction of \$61.10 was processed. [\dots 12 further transactions \dots]''}

\textbf{Prompted Question (Type 6, Variant 1):} \textit{``What's this merchant's total transaction count? Give the exact count.''}

\textbf{Programmatic Derivation:} Count of rows in the structured transaction table $= 14$.

\textbf{Gold Answer Target:} \textit{``This merchant has 14 transactions.''}
\end{quote}

\end{document}